\documentclass[10pt,twocolumn,letterpaper]{article}

\usepackage{wacv}
\usepackage{times}
\usepackage{epsfig}
\usepackage{graphicx}
\usepackage{amsmath}
\usepackage{amssymb}

\usepackage{algorithmic}
\usepackage{multirow}
\usepackage{diagbox}
\usepackage[ruled,linesnumbered]{algorithm2e}
\def\wacvPaperID{115} 

\wacvfinalcopy 

\ifwacvfinal
\fi

\ifwacvfinal
\usepackage[breaklinks=true,bookmarks=false]{hyperref}
\else
\usepackage[pagebackref=true,breaklinks=true,colorlinks,bookmarks=false]{hyperref}
\fi

\ifwacvfinal
\fi

\begin{document}
	
	\title{Compensation Tracker: Reprocessing Lost Object for Multi-Object Tracking}
	
	\author{Zhibo Zou\thanks{These authors contribute equally.} 
		\and Junjie Huang\textsuperscript{*} 
		\\
		School of Automation, Chongqing University of Posts and \\Telecommunications, Chongqing 400065, China\\
		{\tt\small {$\{$s190331072, s190331071$\}$}@stu.cqupt.edu.cn}\\
		{\tt\small {$\{$luoping$\}$@cqupt.edu.cn}}
		\and Ping Luo\\
	}

	\maketitle
	\ifwacvfinal
	\thispagestyle{empty}
	\fi
	
	\begin{abstract}
		Tracking by detection paradigm is one of the most popular object tracking methods. However, it is very dependent on the performance of the detector. When the detector has a behavior of missing detection, the tracking result will be directly affected. In this paper, we analyze the phenomenon of the lost tracking object in real-time tracking model on MOT2020 dataset. Based on simple and traditional methods, we propose a compensation tracker to further alleviate the lost tracking problem caused by missing detection. It consists of a motion compensation module and an object selection module. The proposed method not only can re-track missing tracking objects from lost objects, but also does not require additional networks so as to maintain speed-accuracy trade-off of the real-time model. Our method only needs to be embedded into the tracker to work without re-training the network. Experiments show that the compensation tracker can efficaciously improve the performance of the model and reduce identity switches. With limited costs, the compensation tracker successfully enhances the baseline tracking performance by a large margin and reaches 66\% of MOTA and 67\% of IDF1 on MOT2020 dataset.
	\end{abstract}
	
	\begin{figure*}[ht]
		\begin{center}
			\centerline{\includegraphics[width=2.1\columnwidth]{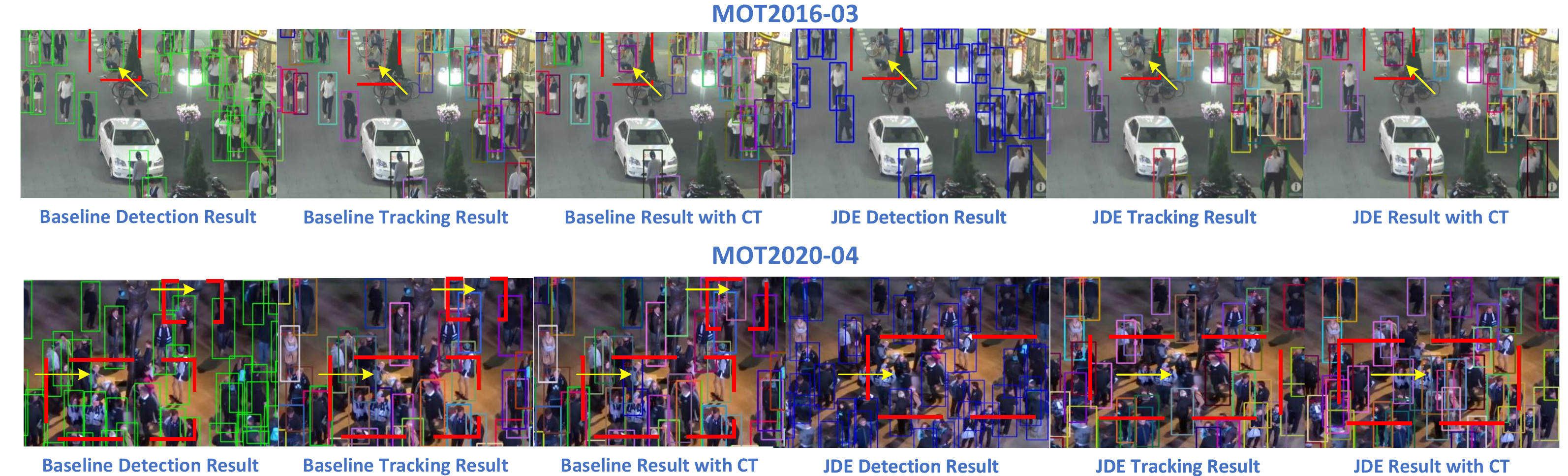}}
			\caption{Comparison diagram of model detection result, model tracking result and model with compensation tracker result. The red dashed boxes is the missing detection area, their tracking result area and compensation area of two models. After motion compensation and object selection, the compensation tracker is able to compensate for some previously tracked objects with high reliability (Yellow Arrow Objects).}
			\label{wacv-fig:1}
			\vskip -0.45in
		\end{center}
	\end{figure*}
	\section{Introduction}
	
	Currently, multi-object tracking is one of the basic tasks of computer vision with wide applications in many fields, for example, intelligent security, automatic driving, pedestrian tracking, intelligent monitoring and so on \cite{ciaparrone2020deep}. Tracking by detection (TBD) framework is the current mainstream tracking paradigm.
	
	In order to improve tracking performance, existing researches usually use multiple sub-networks for object information extraction and data association. Such methods generally use networks or modules with information sharing function such as RNN \cite{milan2017online}, LSTM \cite{kim2018multi} and Siamese network \cite{shuai2020multi,chu2020dasot} to save information of tracked objects and complete spatio-temporal matching. In addition, the use of various object information, for example motion information, spatio-temporal information, appearance information and affinity information \cite{yin2020unified,huang2020refinements,xu2019spatial,chu2019famnet,sun2019deep}, will complicate the information extraction network. Also, these methods will increase additional computation cost and do harm to the inference speed. Many researchers notice that the speed of tracking model is as important as the tracking performance. For the real-time model, the structure of the detection network cannot be too complicated. Simultaneously, methods of data association should be simple and effective. So, real-time model usually uses detector with speed-accuracy trade-off, for instance, RetinaNet \cite{lu2020retinatrack}, CenterNet \cite{zhang2020fairmot,zhou2020tracking,zhou2019objects,zhan2020simple}, Yolov3 \cite{redmon2018yolov3,wojke2017simple,wang2019towards}, Faster RCNN \cite{bergmann2019tracking} and so on. On the other hand, a series of simple and fast data association methods are also proposed such as bounding box (BBox) intersection ratio (IoU) matching \cite{bochinski2017high,bochinski2018extending,bewley2016simple}, motion prediction \cite{kalman1960new,huang2020refinements} and appearance matching \cite{yu2016poi,zhang2020fairmot,yoon2021online,ristani2018features}. This kinds of information can be obtained by traditional methods or can be learned together with detection task. Therefore, these easily available object characteristics are often used for data association in the real-time model.
	
	In the existing real-time model, both end-to-end model and non-end-to-end model are essentially TBD paradigm. In other words, the performance of the detector plays a decisive role in the tracking result. It’s seen in Fig.\ref{wacv-fig:1}. When an object can be tracked in the past frame but cannot be detected in a certain frame due to the unstable performance of the detector (Yellow Arrow Object), the object will not be able to be matched by data association and is considered as an unmatched tracked object. To this end, the tracking result will not include this object and mark them as lost object. This phenomenon will lead to missing tracking and cause ID switches of the missing tracking object so as to impair the tracking performance.
	
	The above discussion raises some questions: \emph{Is the detector stable enough to avoid missing detection? Are these missing tracking objects not in the tracking area or invisible? The answer is negative!}
	In the existing researches, additional neural networks such as Siamese networks \cite{chu2019famnet,yin2020unified,chu2019online} and RNN \cite{fang2018recurrent,xu2019spatial,sadeghian2017tracking} are often used in the model to re-track lost objects. However, these methods will increase a lot of computation costs and are not friendly to real-time tracking. Using lightweight networks to re-track lost objects will not perform well under complicated scenes \cite{chen2018real}. We hope to maintain the speed-accuracy trade-off and re-track the missing tracking objects from the lost objects with computation costs as little as possible. We believe that the tracker not only can exploit the information provided by the detector to perform data association, but also can use the past information to predict positon and compensate for missing tracking objects. Based on this motivation, we propose compensation tracker (CT) based on simple and effective methods to reprocess the lost objects. It contains a motion compensation module and an object selection module. CT screens out the highly credible missing tracking objects from lost objects and re-tracks them without adding abundant computation consumption. As can be seen in Fig.\ref{wacv-fig:1}. The proposed method can effectively alleviate the problem of missing tracking caused by the instability of the detector. Especially in dense crowd regions, the proposed method can greatly improve the tracking performance of the model and reduce ID switches without re-training the network. The contributions of this paper are summarized as follows:
	
	(1)	Indicating and analyzing the lost tracking phenomenon of each sub-dataset on MOT2020 dataset.
	
	(2)	We propose a compensation tracker based on motion compensation and object selection. By performing motion prediction and reasonable features judgment for the lost objects, CT selects out missing tracking objects with high reliability and re-tracks them in the result.
	
	(3) Extensive experiments show that the reasonable use of traditional methods not only can achieve outstanding compensation result, but also only limited computation costs is produced. Particularly in the dense crowd areas, our method has prominent improvements for the tracking performance of the model.
				\begin{table*}
		\begin{center}
			\fontsize{9.5}{9}\selectfont
			\setlength\tabcolsep{2pt}
			\begin{tabular}{|c| c| c| c| c| }
				\hline
				Model&
				\multicolumn{2}{c|}{FairV1}&
				\multicolumn{2}{c|}{JDE}\\
				\hline
				\diagbox[width=5em,trim=l]{Sets}{Objects}&Lost Objects Number &Compensation Objects Number&Lost Objects Number &Compensation Objects Number\\
				\hline
				MOT20-01&5016&1559&5888&1832\\
				MOT20-02&34434&11361&41949&15147\\	
				MOT20-03&86814&32531&72649&30985\\
				MOT20-05&255953&96167&223091&90558\\
				\hline

			\end{tabular}
			\caption{The Statistical number of lost objects and compensation objects in FairV1 and JDE on MOT2020 training dataset.}
			\label{wacv-table:1}
		\end{center}
		
		\vskip -0.25in
	\end{table*}
	\section{Related Works}
	
	\textbf{Towards Real-Time Tracking.} According to the association way of detection information, TBD paradigm can be divided into end-to-end tracking model and non-end-to-end tracking model. POI \cite{yu2016poi} obtains appearance features through GoogLeNet \cite{szegedy2016rethinking} and combines offline and online tracking methods for object similarity association. Similarly, Sort \cite{bewley2016simple} utilizes IoU correlation between the predicted BBox and the detected BBox to accomplish data association. IoU Tracker \cite{bochinski2017high} only employs the IoU between tracked BBox and the detected BBox for data association. Although its tracking speed is very fast, the tracking performance is not accurate enough. Based on Sort, Deep-Sort \cite{wojke2017simple} treats the detection result as the tracking benchmark and uses a Re-Identification network (Re-ID) to acquire appearance information for further data matching. It also proposes a cascade matching method and combines with the Hungarian \cite{kuhn1955hungarian} algorithm for appearance features matching. Subsequently, MOTDT \cite{chen2018real} uses a lightweight network (SqueezeNet\cite{iandola2016squeezenet}) to estimate the trajectory scores of each lost object and makes use of the score map to re-track the missing tracking objects from lost objects. JDE \cite{wang2019towards} simultaneously learns detection information and appearance embedding information in Yolov3 \cite{redmon2018yolov3}. Some researches \cite{zhou2020tracking,bergmann2019tracking,lu2020retinatrack,pang2020quasi} also use other detectors or detection optimization methods to improve tracking performance. Based on the JDE, FairMOTV1 \cite{zhan2020simple} utilizes CenterNet \cite{zhou2019objects} to learn detection information and appearance information. It exploits the advantages of heatmaps detection to obtain object information that is more friendly to data association. FairMOVT2 \cite{zhang2020fairmot} optimizes the training methods and corrects the ground truth (GT) labels on the basis of FairMOTV1. We use FairMOTV1 (FairV1) as the baseline and analyze the phenomenon of lost object on it.

	In the sequels, the pipeline end-to-end methods \cite{sun2020simultaneous,pang2020tubetk} associate tracked objects in spatio-temporal sequences in adjacent frame. The author designs a pipeline IoU learning method and a pipeline IoU matching method for object association. Chained-Tracker \cite{peng2020chained} considers that the BBox of adjacent frame are the paired-matching relationship. The author designs a paired boxes loss and corresponding matching method for data association. CenterTrack \cite{zhou2020tracking} provides a more concise tracking framework. It directly exploits the heatmap offsets direction of adjacent frame for data association. Quasi-Dense \cite{pang2020quasi} makes use of Bi-Softmax and Quasi-Dense similarity learning to improve the robustness of Re-ID features. However, whether it is end-to-end or non-end-to-end tracking paradigm, they excessively rely on the detection quality and emerge from the phenomenon of missing tracking.

	\textbf{Towards Lost Object Tracking.} Using a variety of object information to improve matching accuracy is one of the solutions to avoid lost tracking. Therefore, many researchers prone to use RNN, LSTM \cite{fang2018recurrent,zhu2018online,sadeghian2017tracking,kim2018multi} or other methods with information preservation and decision-making to optimize the data association of multiple object information. Besides, employing Siamese network \cite{chu2020dasot,chu2019famnet,chu2017online} for each detected object is also another network solution. Some researchers utilize additional attention networks to extract more powerful appearance features and solve the ambiguous featue problems caused by frequent object interaction and insufficient affinity \cite{yin2020unified,zhu2018online}. In recent years, graph neural networks (GNN) have been gradually used for object tracking. Some researches \cite{zhou2018online,braso2020learning,ma2019deep,li2020graph} exploit  the powerful information transmission capability and graph matching ability of GNN for data matching so as to improve the tracking performance. Multi-hypothesis tracking \cite{blackman2004multiple} generates real trajectory and hypothesis trajectory by Kalman filtering \cite{kalman1960new}. Through clustering and pruning, the algorithm controls the number of hypothetical trajectories and evaluates the credibility of hypothetical trajectories as real trajectories by log-likelihood estimation scores \cite{blackman2004multiple,kim2015multiple}. This method also depends on the performance of the detector and makes insufficient use of the input image features. Based on IoU Tracker, V-IoU \cite{bochinski2018extending} processes the lost objects with a visual single object tracker to enhance the continuity of the trajectory. But it depends on the performance of the filter and it will fail to track the missing tracking object under the condition of object deformation and camera motion. Our method only needs to be embedded into the tracker and does not require extra networks. When the object cannot be matched by the basic tracker, CT firstly performs motion compensation to re-predict its position. Then the motion compensation information and hand-crafted appearance information of the object are used to determine whether the object is credible or not. Our method not only inherits the advantages of V-IoU, but also takes the advantages of appearance information to improve the accuracy of object selection.
	
			\begin{figure}[t]
						\vskip -0.1in
		\begin{center}
			\includegraphics[width=0.88\linewidth]{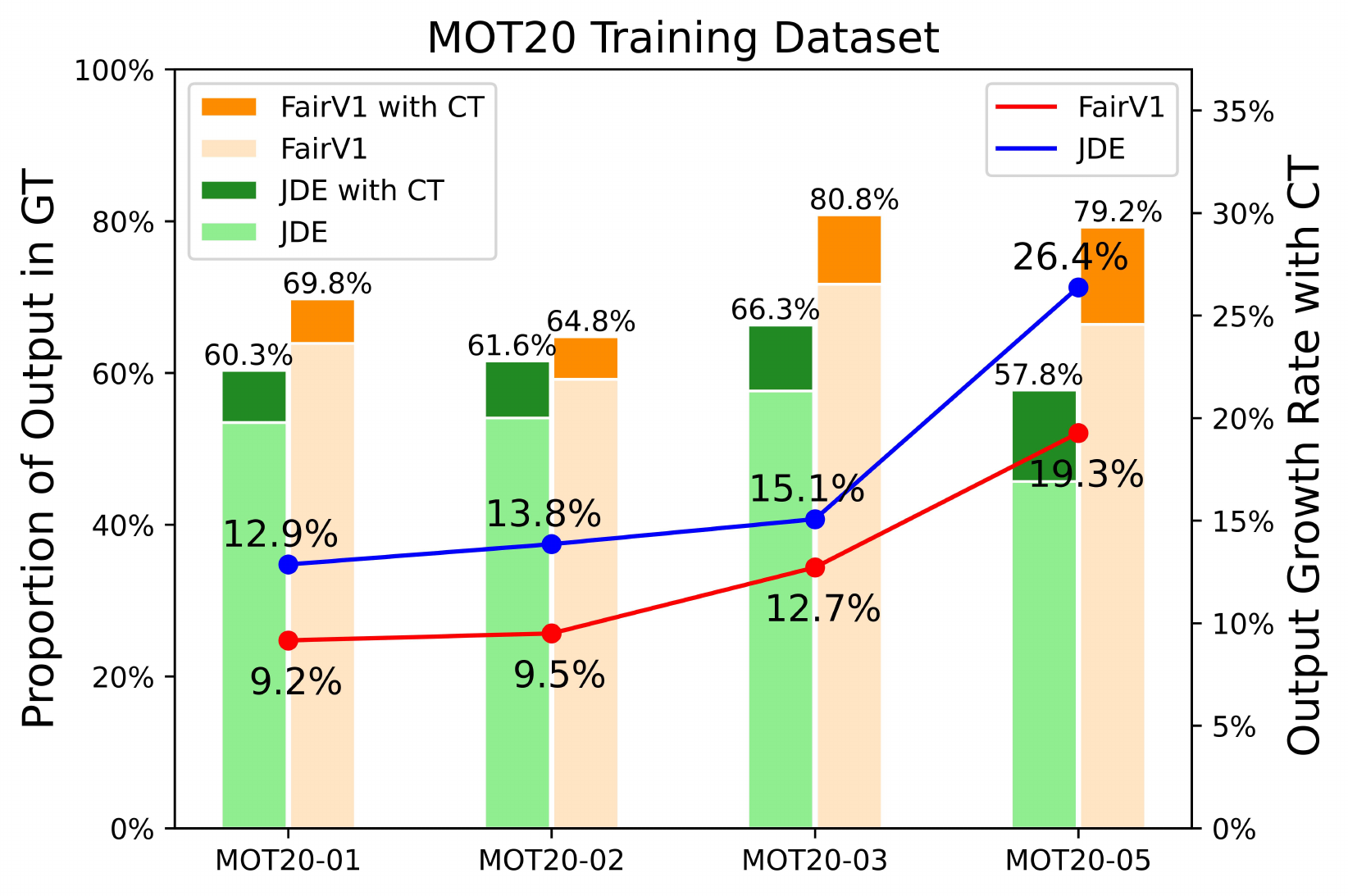}
		\end{center}
		\caption{Statistical result of MOT2020 training dataset. The histogram shows the percentage between the tracking result of each sub-dataset and GT in two models. The line chart shows the tracking result growth rate of two models after embedding CT.}
		\label{wacv-fig:2}
		\vskip -0.2in
	\end{figure}
	\section{Analysis of Lost Object}
	\textbf{Experiments Setting.} All trained network parameters in experiments are from the corresponding authors. The model parameters used for statistic and analysis have not been trained by MOT2020 training dataset. We use DLA34 \cite{yu2018deep} with DCNv2 \cite{zhu2019deformable} in FairV1 and JDE-1088x608 in JDE for statistical experiments. All hyper-parameters will be consistent and the same as authors setting.

	\textbf{Analysis on MOT2020 Result.} MOT2020 dataset is a dense crowd sequences with an average of 246 pedestrians in each frame \cite{dendorfer2020mot20}. These sequences include indoor/outdoor and day/night scenes. The sequences with dense crowds and insufficient illumination are a great challenge for detection stability and tracking performance. As can be seen in Fig.\ref{wacv-fig:2}. The tracking quantity of two models (FairV1 \& JDE) accounts for about 60\% in GT. These results indicate that the generalization ability of two models are not good enough in dense regions. There is a high probability in lost tracking due to the poor performance of detector. As can be seen in Table \ref{wacv-table:1}, the two models have lost objects in each sub-dataset. In lost objects set, there are some missing tracking objects caused by missing detection. We believe that the tracker can make full use of the existing tracked information to predict position and compensate the output for missing tracking objects. Based on this idea, we use the motion compensation module and the object selection module in CT instead of CNN or GNN. The motion compensation module is used to re-predict the position of the lost object while the object selection module is responsible for feature selection and feature matching. By position judgment and hand-crafted feature matching, CT selects out the highly reliable objects (missing tracking objects) from lost objects and compensates output sequences for them (compesation objects). As can be seen in Fig.\ref{wacv-fig:2}. To this end, FairV1 can be added the tracking sequences by an average of 12.7\% and JDE can be increased in the tracking sequences by an average of 17\%. This result proves that our compensation objects is effective and credible. To sum up, CT succeeds in improving the tracking performance of the model in dense crowd areas. 
		\begin{figure*}
		\begin{center}
			\includegraphics[width=1.01\linewidth]{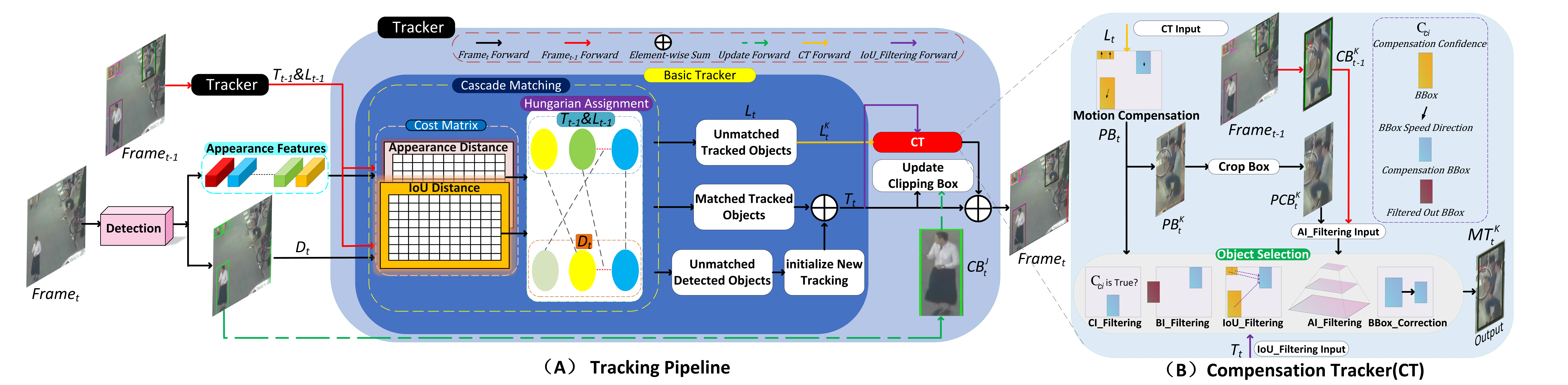}
		\end{center}
		\caption{Online tracking pipeline of compensation tracker. The above figure shows the overall data flow and the relationship between the basic tracker and the compensation tracker. Actually, CT is the extension of the basic tracker. When an object is tracked at frame $t-1$ and is not detected at frame $t$, the object cannot be matched by cascade matching and is considered as lost object $L_t^K$. Then, $L_t^K$ is transmitted in the compensation tracker for motion compensation and object selection. Meanwhile, tracked object $J$ will be updated their information. After processing the selection result and the association result, the tracker outputs the final tracking result for object tracking.}
		\label{wacv-fig:3}
		\vskip -0.2in
	\end{figure*}
	\section{Methods}	
	\subsection{Design of Compensation Tracker}
	In this section, we will introduce the compensation tracker in detail. It contains three main tracking modules: basic tracker, motion compensation (MC) and object selection (OS). The tracking pipeline can be seen in Fig.\ref{wacv-fig:3} ($K$ and $J$ represent the object number in the corresponding set).
	
	\textbf{Basic Tracker.} Given an image of frame $t$, the detected object set $D_t$ can be obtained by network recognition. At frame $t-1$, we can get successfully tracked object set $T_{t-1}$ and lost tracking object set $L_{t-1}$ from the previous frame. Then, the basic tracker performs cascade matching including IoU matching and appearance matching on $D_t$, $L_{t-1}$ and $T_{t-1}$ \cite{zhan2020simple,wang2019towards}. Cascade matching outputs three kinds of objects containing matched tracked objects, newly detected objects (unmatched detected objects) and lost tracking objects (unmatched tracked objects). For the matched tracked objects and the newly detected objects, we update their information and their tracked clipping box images $CB_t$. Subsequently, the matched tracked objects and the newly detected objects are transmitted in the tracked object set $T_t$ (tracking result). And the lost tracking object set $L_t$ is inputed in CT. 
	
	\textbf{Motion Compensation.} Motion compensation module re-predicts the positions for lost tracking object set $L_t$ and outputs the re-predicted results set $PB_t$. Next, CT implements object selection based on re-predicted results $PB_t$. 
	 
	\textbf{Object Selection.} For filtering out the invisible BBox and correcting BBox's size, CT carries out confidence interference filtering, boundary interference filtering and BBox correction for predicted $PB_t$. Afterwards, IoU interference filtering is used for suppressing the overlap BBox between the predicted $PB_t$ and the tracked object set $T_t$. Next, we crop the BBox image $PCB_t^K$ according to the predicted BBox. Because the object is tracked in the previous frame, the finally tracked clipping BBox image $CB_{t-1}^K$ is retained. We use  $PCB_t^K$ and $CB_{t-1}^K$ for hand-crafted appearance extraction and matching in appearance interference filtering. If an object can be selected out by object selection, we consider that it is the missing tracking object $MT_t^K$ that caused by failing to be detected and we compensate for them in the tracked object set $T_t$. Otherwise, this lost object that cannot be compensated will be saved in $L_t$ at most 30 frames.
	
	\subsection{Motion Compensation}
	In the motion compensation module, we use the Kalman filter with uniform motion and linear observation by default. Its input can be defined as:
	\begin{equation}
	{Mean}=[x,y,a,h,\dot{x},\dot{y},\dot{a},\dot{h}]
	\label{eq:eq1}
	\end{equation}
	where $x$ and $y$ are the horizontal and vertical coordinates of the BBox respectively. $a$ is the ratio between BBox's width and BBox's height. $h$ is the height of the BBox. $\dot{x},\dot{y},\dot{a},\dot{h}$ are the velocities of the corresponding components. $[x,y,a,h]$ are directly observed as object states. 
	
	Then, taking $Mean_{t-1}$ as input and calculating the error covariance matrix between the calculation value at frame $t$ and the real value at frame $t-1$:
	
	\begin{equation}
	{{Mean_t^{'}}}=F_tMean_{t-1}+A_tX_t
	\label{eq:eq2}	
	\end{equation}
	\begin{equation}
	{Cova_t^{'}} = F_tCova_{t-1}{F_t}^\mathbb{T}+Q
	\label{eq:eq3}	
	\end{equation}
	where ${{Mean_t^{'}}}$ is the estimation value of the system state at frame $t$. $Mean_{t-1}$ is the real value of the system state at frame $t-1$. $F_t$ is the motion transformation matrix from the previous state to the current state. $A_t$ is the control martix. $X_t$ is the control variable. ${Cova_t}^{'} $ is the covariance matrix of the error between the calculation value and the real value. $Cova_{t-1}$ is the covariance matrix of the error between the estimation value and the real value. $\mathbb{T}$ is the transpose operator. $Q$ is the multi-variate normal distribution of covariance matrix. Next, Kalman gain is calculated:
	\begin{equation}
	K_t={Cova_t}^{'}{H_t}^\mathbb{T}(H_t{Cova_t}^{'}{H_t}^\mathbb{T}+R)^{-1}
	\label{eq:eq4}	
	\end{equation}
	\begin{equation}
	Mean_{t}={Mean_t^{'}}+K_t(Z_t-H_t{Mean_t^{'}})
	\label{eq:eq5}	
	\end{equation}
	where $K_t$ is the Kalman gain. $Z_t$ is the system measurement value. $H_t$ is the measurement transfer matrix and $R$ is covariance matrix of observation noise \cite{wojke2017simple,kalman1960new}.
	
	Finally, the error covariance matrix between the estimation value and the real value is updated:
	\begin{equation}
	Cova_t=(1-K_t\cdot H_t){Cova_t^{'}}
	\label{eq:eq6}	
	\end{equation}
	\subsection{Object Selection}
	Experiments show that only using the motion compensation module to unconditionally compensate for the lost objects will result in wrong compensation. In this section, we will further introduce how to screen the BBox predicted by motion compensation to find out the missing tracking object and compensate for them.
	
\begin{figure}[ht]
		
		\begin{center}
			\centerline{\includegraphics[width=1.0\columnwidth]{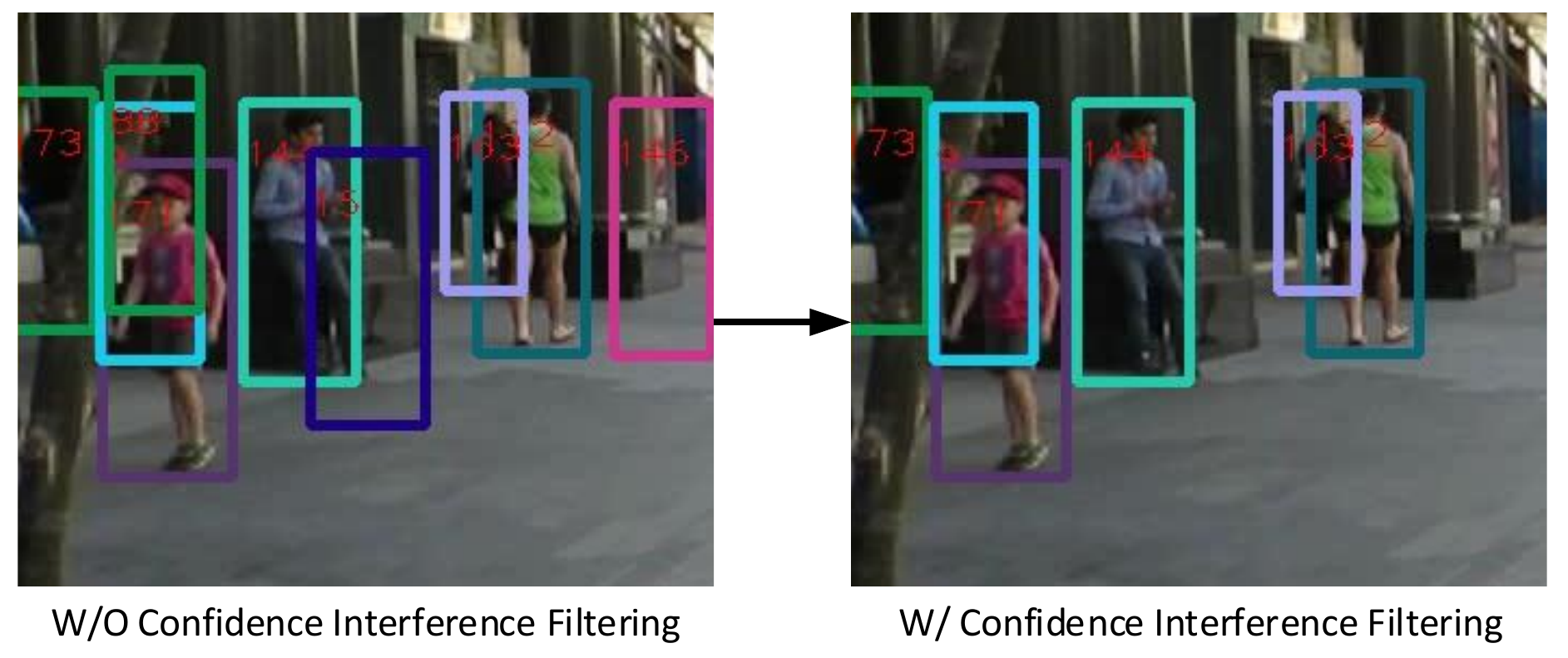}}
			\caption{Diagram of confidence interference filtering result. After using the confidence interference filtering, the BBox (ID 15, ID 98 and ID 146) that does not meet the formula (\ref{eq:eq7}) will be filtered out and the others are retained because they satisfy the condition.}
			\label{wacv-fig:4}
		\end{center}
		\vskip -0.25in
	\end{figure}

	\textbf{Confidence Interference Filtering(CI\_Filtering).} Error bounding box (EBBox) is generated by the fact that the tracked object has been lost or the object has not been detected in so many frames, but the motion compensation is still carried out. Referring to the maximum reserving frame value of the lost object in the baseline, we consider it as the  maximum of compensation frame value and employ the  compensation confidence threshold to avoid the generation of EBBox. The compensation confidence is defined as follow:
	\begin{equation}
	\mathcal{C}_{ci}=\mathbb{I}\left\{S_{ts}-L_{ts}>0\right\} \quad \quad s.t. S_{ts} >C_F
	\label{eq:eq7}	
	\end{equation}
	where $\mathcal{C}_{ci}$ is the compensation confidence. $S_{ts}$ is the number of times that the object is successfully tracked. $L_{ts}$ is the number of times that the object is lost in tracking and $C_F$ is the compensation frame value. When a lost object does not meet the above formula, the object will be filtered out. It’s seen in Fig.\ref{wacv-fig:4}.

\begin{figure}[ht]
		\vskip -0.1in
		\begin{center}
			\centerline{\includegraphics[width=1.0\columnwidth]{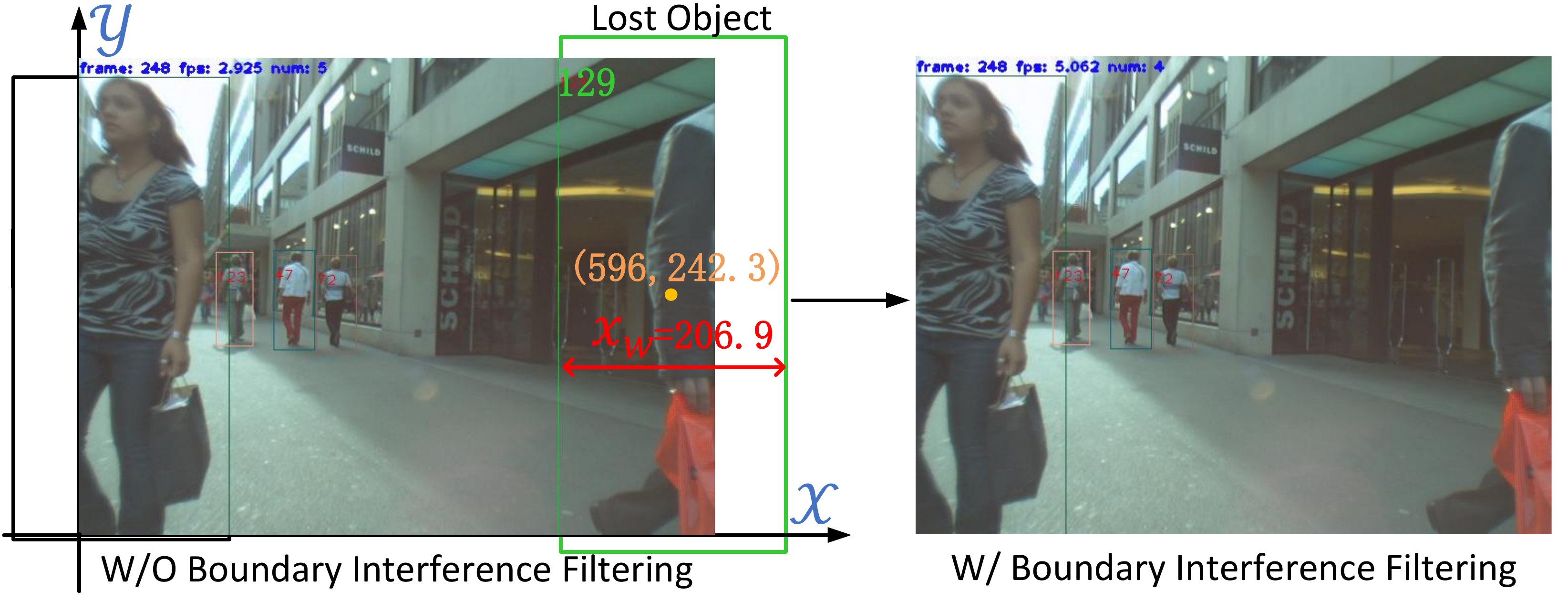}}
			\caption{Diagram of boundary interference filtering result. The width and height of the image are 640 and 480 respectively. The center coordinate of the lost object 129 is (596, 242.3) and its width $x_w$ is 206.9. Because the object 129 does not satisfy the formula (\ref{eq:eq8}), it will be filtered out.}
			\label{wacv-fig:5}
		\end{center}
		\vskip -0.25in
	\end{figure}

	\textbf{Boundary Interference Filtering(BI\_Filtering).} When the tracked scene moves relatively fast, only using the confidence interference filtering will not achieve an optimal compensation result. When a lost object moves away from the tracking area, we need to judge the center position of its predicted BBox by the following formula:
	\begin{equation}
	\mathcal{C}_{bi} = \mathbb{I}\left\{x-x_w *\alpha >0 \quad
	\land \quad  w-x-x_w * \alpha >0\right\}
	\label{eq:eq8}
	\end{equation}
	where $\mathcal{C}_{bi}$ is the confidence of BI\_Filtering. $x$ is the center point abscissa of the predicted BBox. $x_w$ is the width of the predicted BBox. $\alpha$ is a boundary weight. $w$ is the width of the image. When a predicted BBox of lost object does not satisfy formula (\ref{eq:eq8}), the object will be filtered out. As it's seen in Fig.\ref{wacv-fig:5}.

\begin{figure}[ht]
			\vskip -0.05in
		\begin{center}
			\centerline{\includegraphics[width=0.7\columnwidth]{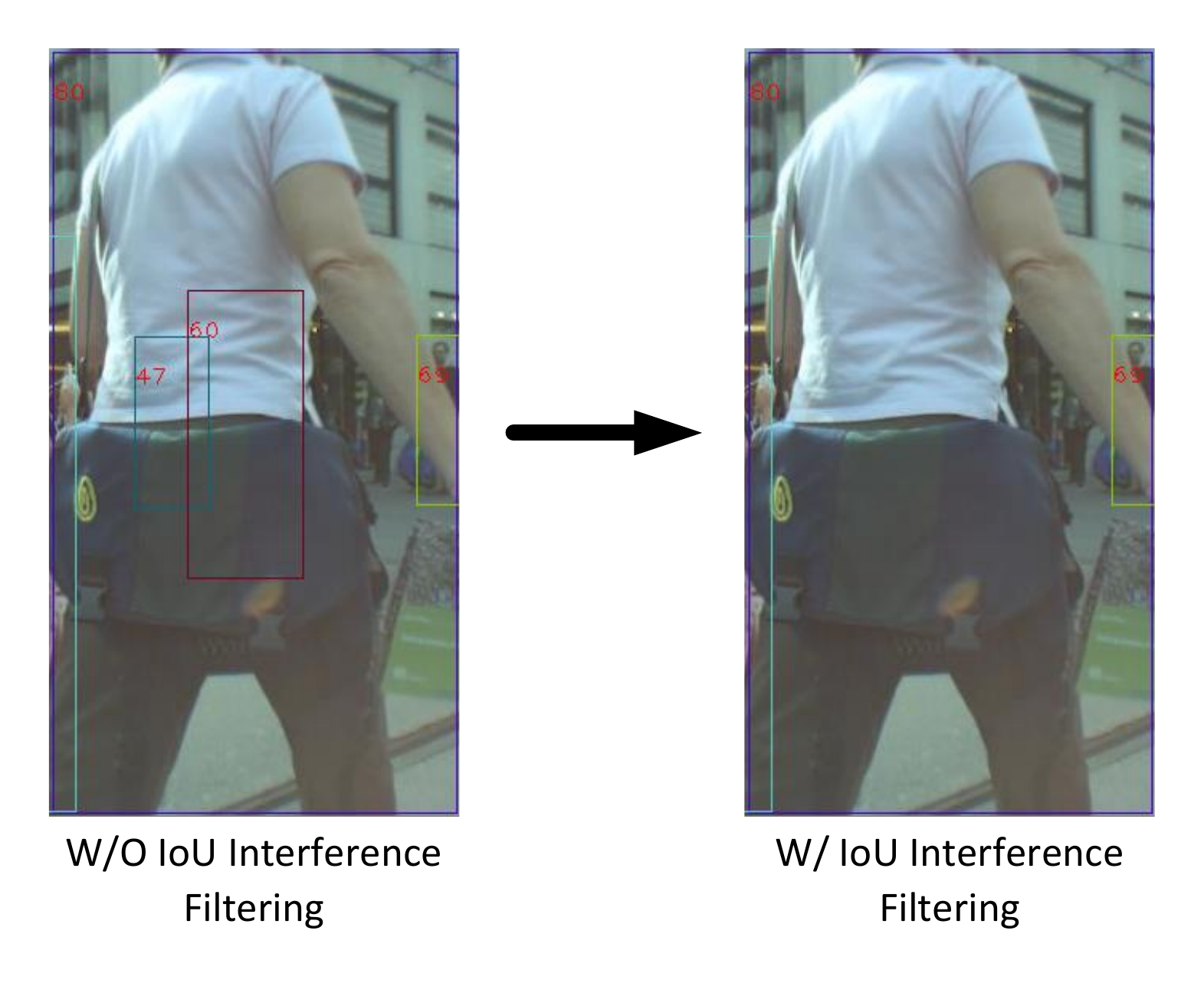}}
			\caption{Diagram of IoU interference filtering result. It is obvious from the figure that the wrong BBox has been removed.}
			\label{wacv-fig:6}
		\end{center}
		\vskip -0.4in
	\end{figure}

	\textbf{IoU Interference Filtering(IoU\_Filtering).} In order to improve the compensation result, we also eliminate the BBox with object occlusion and object overlap to further prevent wrong compensation. The filtering effect can be seen in Fig.\ref{wacv-fig:6}. We compare the predicted compensation BBox set $PB_t$ with the tracked object set $T_t$ containing their area ratio, IoU and BBox embedding degree.

	\textbf{Appearance Interference Filtering(AI\_Filtering).} IoU Filtering cannot solve the problem of occlusion by other objects and severe drift of BBox. So, we use the position information predicted by motion compensation to extract the clipping box image $PCB_t^K$ at the current frame. Afterwards, the hand-crafted feature similarity between $PCB_t^K$ and the finally tracked clopping box image $CB_{t-1}^K$ will be evaluated. Firstly, the two clipping BBox images will be filtered by Gaussian kernel and down-sampled twice respectively. Then, the Gaussian difference calculation between them and themselves will be performed \cite{lowe2004distinctive}:
	\begin{equation}
	D(p_x,p_y,\sigma) = L(p_x,p_y,S_f\sigma)-L(p_x,p_y,\sigma)
	\label{eq:eq9}
	\end{equation}
	where $D(\cdot)$ is a Gaussian differential operator. $D(p_x,p_y,\sigma)$ is the Gaussian differential image. $p_x$ and $p_y$ are pixel position of the image. $L(\cdot)$ is the Gassian smoothed image which is produced from the convoluition of a variable-scale Gassian with an input image. $\sigma$ is the scale space operator of the Gauss kernel function. $S_f$ is the feature pyramid scaling factor. After getting $D(p_x,p_y,\sigma)$ from formula (\ref{eq:eq9}), the algorithm finds the extremum in it and inputs the extremum point sets into formula (\ref{eq:eq10}) to obtain the modulus and direction of each extremum region \cite{lowe2004distinctive}.
	\begin{equation}
	\begin{cases}
	\begin{split}
	m(p_x,p_y)=||(L(p_{x+1},p_y)-L(p_{x-1},p_y))+\\
	(L(p_x,p_{y+1})-L(p_x,p_{y-1}))||_{2}
	\end{split}
	\\
	\theta(p_x,p_y) =tan^{-1}(\frac{L(p_x,p_{y+1})-L(p_x,p_{y-1})}{L(p_{x+1},p_y)-L(p_{x-1},p_y)})
	\end{cases}
	\label{eq:eq10}
	\end{equation}
	where $m(p_x,p_y)$ is the modulus of the extreme points and $\theta(p_x,p_y)$ is the direction of the extreme points.
	
	Set $m$ and set $\theta$ are input into the Histogram of Oriented Gradients \cite{dalal2005histograms} to generate the 128 dimension vectors set. Finally, K-Nearest-Neighbor (KNN) matching is performed for measuring the Euclidean distance of these vectors  \cite{fukunaga1975branch}. If there are enough matching points, the lost object is treated as the missing tracking object and will be re-tracked. The matching formula is as follows:
	\begin{equation}
	\mathcal{C}_{ai}=\mathbb{I}\left[\mathcal{K}(\mathcal{H}(m_{t-1}^K,\theta_{t-1}^K),\mathcal{H}(m_{t}^K,\theta_t^K))>\sigma_m\right]
	\label{eq:eq11}
	\end{equation}
	where $\mathcal{C}_{ai}$ is the confidence of AI\_Filtering. $\mathcal{K}$ is the KNN matching function. $\mathcal{H}$ is the Histogram of Oriented Gradients. $m_{t-1}^K$ and $\theta_{t-1}^K$ are the modulus and direction of $CB_{t-1}^K$ respectively. $m_t^K$ and $\theta_t^K$ are the modulus and direction of $PCB_t^K$ respectively. $\sigma_m$ is the threshold for KNN matching.
	
	\textbf{BBox Correction(BBox\_Correction).} Since the predicted BBox will have an inaccurate size, the object cannot be marked accurately. The BBox's size in adjacent frame changes very limited. When the change of area between lost BBox and its predicted BBox is greater than 1.1, the compensation BBox will be resized in the original size. Algorithm \ref{algu:1} describes how the proposed method works for the lost objects.
	\begin{algorithm}[tb]
		\vskip -0.in 
	\caption{Compensation Tracker}
	\label{algu:1}
	\KwIn{\\
		$L_t$ is the lost object set at frame $t$\\
		$T_t$ is the tracked object set at frame $t$\\
		${CB}_{t-1}^K$ is the K-th clipping box image at frame $t-1$\\
		${PCB}_{t}^K$ is the K-th clipping box image at frame $t$}
	
	\KwOut{\\
		$MT_t$ is the missing tracking object set at frame $t$}
	\begin{algorithmic}[1]
		\FOR{$\l_K \in L_t$}
				\STATE ${Mean_t},{Cova_t}$ $\leftarrow$ MC($l_K.Mean$,
		$l_K.Cova$)      by(\ref{eq:eq1})(\ref{eq:eq2})(\ref{eq:eq3})(\ref{eq:eq4})(\ref{eq:eq5})(\ref{eq:eq6}) 
				\STATE $Mean_t$  $\leftarrow$ BBox$\_$Correction ($Mean_t$,$l_K.Mean$)
		\STATE $\mathcal{C}_{ci}$ $\leftarrow$ CI$\_$Filtering($l_K$) by(\ref{eq:eq7})
		\STATE $\mathcal{C}_{bi}$ $\leftarrow$ BI$\_$Filtering ($Mean_t$) by(\ref{eq:eq8})
				\FOR{$T_t^l$ $\in$ $T_t$}
		\STATE $l_K$ $\leftarrow$ IoU$\_$Filtering($Mean_t$,$T_t^l$)
		\ENDFOR
		\STATE $\mathcal{C}_{ai}$ $\leftarrow$ AI$\_$Filtering (${CB}_{t-1}^{K}$,${PCB}_t^{K}$) \\
		 by(\ref{eq:eq9})(\ref{eq:eq10})(\ref{eq:eq11})
		\IF{$\mathcal{C}_{ci} \land \mathcal{C}_{bi}$ $\land$ $\mathcal{C}_{ai}$} 
		\STATE  $l_K$ $\leftarrow$ Update$\_$Parameter($Mean_t,Cova_t$)
		\STATE $MT_t.append(l_K)$ 
		\ENDIF
		\ENDFOR
		\STATE \textbf{return} $MT_t$
	\end{algorithmic}
\end{algorithm}

	\section{Experiments}
	\subsection{Experiment Details }
	We carry out extensive evaluations on MOT2015\cite{leal2015motchallenge} dataset, MOT2016\cite{milan2016mot16} dataset, MOT2017\cite{milan2016mot16} dataset and the latest MOT2020\cite{dendorfer2020mot20} dataset. Besides, we achieved relatively good results and metrics precision in all datasets.
	
	\textbf{Experiments Platform.} Our experiments are implemented on Pytorch. The computer used in the experiment is equipped with Xeon Platinum 8163 CPU and RTX2080Ti. Furthermore, we use the trained model parameters of the DLA network for experiments in baseline. And JDE-1088x608 is employed in JDE in ablation experiments. All the trained model parameters are provided by corresponding author. These networks are trained with the following datasets including MOT2017\cite{milan2016mot16}, Caltech\cite{dollar2009pedestrian}, CityPersons\cite{zhang2017citypersons}, CuhkSysu\cite{xiao2017joint}, PRW\cite{zheng2017person} and ETH\cite{ess2008mobile}.
	
	\textbf{Evaluation Metrics.} CLEAR MOT Metrics \cite{bernardin2008evaluating} and IDF1 are used for measuring tracking result. These metrics include multi-object tracking accuracy (MOTA), ID switches (ID.Sw), the ratio of correctly identified detections over the average number of GT and computed detections (IDF1), multi-object tracking precision (MOTP), the most tracked object (MT), the most lost object (ML), the average number of false alarms per frame (FAF), the total number of times a trajectory is fragmented (Frag), the number of false positives (FP) and the number of false negatives (FN) \cite{ciaparrone2020deep}. And FPS is obtained by running on a single RTX2080Ti. 
	\begin{table}
		\begin{center}
			\vskip -0.2in
				\setlength\tabcolsep{3.5pt}
				\fontsize{7}{10}\selectfont
			\begin{tabular}{c|c c c |c| c c c }
				\hline
				$C_F$ &MOTA$\uparrow$&IDF1$\uparrow$&ID.SW$\downarrow$&$\sigma_m$ &MOTA$\uparrow$&IDF1$\uparrow$&ID.SW$\downarrow$\\
				\hline
				10 &{\color{red}{46.5}}/76.0 &66.0/76.8&284/516&1&45.5/{\color{red}{76.1}}&65.2/76.7&278/508\\
				20&46.2/76.0&{\color{red}{66.3}}/76.9&266/514&3&45.7/76.0&64.9/{\color{red}{76.9}}&284/514\\
				30&46.2/{\color{red}{76.0}}&65.6/{\color{red}{77.1}}&{\color{red}{263}}/518&5&46.0/76.0&{\color{red}{65.5}}/76.8&282/509\\
				40&46.0/76.0&65.8/76.8&269/{\color{red}{510}}&7&{\color{red}{46.2}}/76.0&64.8/76.8&{\color{red}{274}}/{\color{red}{506}}\\
				\hline
			\end{tabular}
			\setlength\tabcolsep{6.5pt}
			\vskip 0.1in
			\begin{tabular}{c|c c c c c }
				\hline
				$\alpha$ &MOTA$\uparrow$&IDF1$\uparrow$&ID.SW$\downarrow$&FP$\downarrow$&FN$\downarrow$\\
				\hline
				0.1&46.1/76.0&65.5/76.8&249/508&18940/7467&{\color{red}{4061}}/18534\\
				0.3&46.2/76.0&65.5/{\color{red}{77.0}}&{\color{red}{249}}/511&18891/7449&4065/{\color{red}{18526}}\\
				0.5&{\color{red}{46.2}}/{\color{red}{76.0}}&{\color{red}{65.5}}/76.9&250/{\color{red}{505}}&18871/{\color{red}{7434}}&4071/18542\\
				0.7&46.2/76.0&65.5/76.8&253/512&{\color{red}{18855}}/7436&4083/18565\\
				\hline
				
			\end{tabular}
			\caption{Hyper-parameters experiments on MOT15/MOT16. We uses $C_F$=30, $\sigma_m$=5 and $\alpha$=0.5 for all datasets and experiments.}
			\label{wacv-table:2}
		\end{center}
		\vskip -0.3in
	\end{table}
	\subsection{Hyper-Parameter Experiments}
	There are three adjustable parameters in CT. They are the compensation frame value $C_F$, matching threshold $\sigma_m$ and boundary weight $\alpha$ respectively. Parameters setting do not require heuristic algorithms and do not need to be adjusted for each dataset. $C_F$ is a hyper-parameter that already exists in baseline and JDE. Considering that it also has an impact on the compensation result, we also discuss its robustness. The sensitivity experiments of three hyperparameters on MOT15 and MOT16 are shown in Table \ref{wacv-table:2}. $C_F$ is the maximum pre-stored threshold for lost objects. Too large $C_F$ will result in too many pre-stored lost objects and too many memory costs. This setting will lead to ambiguous object matching and reduce tracking accuracy. Too small $C_F$ will raise the problem that missing tracking objects cannot be re-tracked and its ID switches frequently. $\sigma_m$ affects the appearance matching precision of missing tracking objects. If $\sigma_m$ is too large, the objects with high reliability cannot be re-tracked. Besides, $\alpha$ is used to filter out the objects that are out of boundary and deal with the drifting problem of the predicted BBox. It only has weak influences on IDF1, FP and FN. Experiments on MOT15 and MOT16 show that the hyper-parameters in CT are robust.
	\begin{table}[t]
		\setlength\tabcolsep{0.75pt}
		\vskip -0.20in
		\begin{center}
			\begin{small}
				\begin{tabular}{c|c c c c c c}
					\hline		
					\multicolumn{6}{c}{MOT2020 Test Dataset}\\
					\hline
					Component & MOTA$\uparrow$ & IDF1$\uparrow$ & MT$\uparrow$ & ML$\downarrow$ & ID.Sw$\downarrow$ & FPS $\uparrow$\\
					\hline
					Baseline & 58.7& 63.7&{\color{red}66.3$\%$}&{\color{red}8.5$\%$}&6013& \color{red}{15.2} \\ 
					Baseline+MC & 65.0&66.6&59.1$\%$&13.0$\%$&    {\color{red}2119}& 14.6 \\
					Baseline+MC+OS &{\color{red}66.0}&{\color{red}67.0}&56.3$\%$&13.3$\%$&2237 &13.5 \\
					\hline
				\end{tabular}
				
			\end{small}
			\caption{Module ablation experiments on MOT2020.}
			\label{wacv-table:3}
		\end{center}
		\vskip -0.30in
	\end{table}
	
	\begin{table}[t]
		\setlength\tabcolsep{0.5pt}
		\vskip 0.1in
		\begin{center}
			\begin{small}
				
				\begin{tabular}{c|c c c c c c}
					\hline
					Models & MOTA$\uparrow$ & IDF1$\uparrow$ & MT$\uparrow$ & ML$\downarrow$ & ID.Sw$\downarrow$&FPS$\uparrow$\\
					\hline
					\multicolumn{7}{c}{MOT2016 Test Dataset}\\
					\hline		
					JDE\cite{wang2019towards} & 64.4& 55.8&35.4$\%$&20.0$\%$&1544&\color{red}{22.3} \\ 
					JDE with CT & {\color{red}65.0} 
					&{\color{red}59.1}
					&{\color{red}36.1$\%$}
					&{\color{red}18.8$\%$}
					&{\color{red}1525}&19.7	\\
					\hline
					FairV1\cite{zhan2020simple} &68.7
					&70.4
					&39.5$\%$
					&19.0$\%$
					&953&\color{red}{21.7}\\
					FairV1 with CT & {\color{red}69.8} & {\color{red}71.1} &{\color{red}42.0$\%$}&{ \color{red}15.8$\%$} &{\color{red}912}&19.2
					\\
					\hline
					\multicolumn{7}{c}{MOT2020 Training Dataset}\\
					
					\hline
					JDE*\cite{wang2019towards}& 48.2
					& 32.1
					& 318
					& 497
					& 18631&\color{red}{15.0}\\
					JDE with CT&{\color{red}54.4}
					&{\color{red}43.1}
					&{\color{red}526}
					&{\color{red}372}
					&{\color{red}11157}&12.3\\
					\hline
					FairV1\cite{zhan2020simple}*&62.3
					&47.5
					&790
					&288
					&16395&\color{red}{15.2}\\
					FairV1 with CT&{\color{red}65.6}
					&{\color{red}57.5}
					&{\color{red}1030}
					&{\color{red}247}
					&{\color{red}7816}&13.5\\
					\hline
					
				\end{tabular}
				
			\end{small}
			\caption{'Private' model ablation experiments. '*' means that the result is evaluated by ourselves.}
			\label{wacv-table:4}
		\end{center}
		\vskip -0.3in
	\end{table}
	
	\subsection{Ablation Experiments}
	
	As can be seen in Table \ref{wacv-table:3}. After using the motion compensation module, some metrics such as MOTA, IDF1 and ID.Sw have been significantly improved due to compensation for the missing tracking objects. However, because of unconditional compensation, some EBBoxes still appear. After conducting the object selection, MOTA and IDF1 can be further improved while only a little quantity of ID switches are raised. Additionally, the speed costs of motion compensation and object selection are only 0.6 FPS and 1.1 FPS respectively.
	
	As can be seen in Table \ref{wacv-table:4}. On MOT2016 test dataset, the result of the JDE with CT is better than itself. CT improves JDE by 3.3\% on IDF1, 0.6\% on MOTA, 0.7\% on MT and 1.2\% on ML. This result manifests that our tracker can effectively enhance tracking performance and optimize data association of lost objects. Especially on the 2020 training dataset, the improvement is more prominent and various metrics achieve greater gains. Among this metrics, MOTA increased by 6.2\%, IDF1 increased by 11\% and ID.Sw decreased by 7474. For the baseline model, various metrics have also been improved on MOT2016 dataset. CT ameliorates baseline by 1.1\% on MOTA, 2.5\% on MT, 3.2\% on ML and 41 on ID.Sw. It can also be seen in Table \ref{wacv-table:4} that there is a significant gains in MOT2020 tracking result. After employing CT, tracking instability problem is further alleviated and various metrics such as MOTA, IDF1 have been improved to a certain extent. More importantly, ID.Sw on MOT2020 surpasses baseline by 8579.
	
	This is because CT alleviates the lost tracking problem caused by the detector instability so that missing tracking objects can be effectively re-tracked. Also, our tracker not only accurately compensates for missing tracking objects, but also lessen unnecessary ID switches. Something is worthy noticeable that time consumption of CT is limited and the average costs approximate 2-3 frames on two 'Private' real-time models.
	
	For the sake of further demonstrating the data association performance of the CT, Sort, MOTDT and Tracktor ('Public' Tracker) are also used for 'Pubilc' ablation experiments.
	
	As can be seen in Table \ref{wacv-table:5}. After using CT, the performance of Sort can be further improved on MOT2020 test dataset. Among these metrics, MOTA increased by 0.6\% and MT increased by 0.9\%. Especially in ID.Sw and Frag, the reductions are 1499 and 10313 respectively. On MOT2020 training dataset, Sort with CT can obtain 7.1 gains, 18.1 gains to Sort counterparts in MOTA and IDF1. Above all, ID.Sw declines by 9253. It can be seen in Table \ref{wacv-table:5} that the tracking continuity (MT \& ML) have also been greatly improved. On the other hand, MOTDT utilizes SqueezeNet \cite{iandola2016squeezenet} to score and re-track lost objects. We replace this network with CT. As it’s seen in Table \ref{wacv-table:5}. There is a great challenge for the performance of the lightweight network on MOT2020 dataset. However, CT uses historical reliable information and hand-crafted featrue information obtained by traditional methods. The extraction way of traditional methods is stabler than network prediction under the complicated circumstances. Compared with MOTDT, our method has various degrees improvements on MOT2020 dataset. Notably on MOT2020 test dataset, our method achieves 2513 gains in Frag, 211 gains in ID.Sw, 1.4 gains in MOTA, 4.4\% gains in MT and 1.1 gains in IDF1. Also, we replace Tracktor's camera motion wtih CT for experiments on MOT2020. Although CT's ID.Sw is higher than camera motion, CT still outperforms Tracktor on MOTA, MT and ML. There are 2.3 gains and 1.2 gains in MOTA of Tracktor wtih CT on MOT2020 training dataset and test dataset respectively. In terms of the performance in 'Public' ablation experiments, CT is effective in data association and has the capability to re-track missing tracking objects.
	
	\subsection{Comparison with state-of-art Online Models}
	\textbf{MOT2016 \& MOT2017.} As can be seen in Table \ref{wacv-table:6}, the evaluation result of our method on MOT2016 and MOT2017 are outstanding. There are improvements in MOTA (69.8\%), IDF1 (71.1\%), MT (42\%) and ML (15.8\%) on MOT2016 test channel. On MOT2017 test channel, MOTA, IDF1, MT and ML are 68.8\%, 70.2\%, 40.8\% and 17.7\% respectively. These metrics are higher than other online traking models and reach the state-of-art performance.
	
	\textbf{MOT2020.} The effect of CT is more remarkable on MOT2020 test result. It can be clearly seen in Table \ref{wacv-table:7}. CT can reach great gains in MOTA (66\%), IDF1 (67\%), Frag (4154) and ID.Sw (2237). What's more, CT surpasses FairV1 by 7.3\% on MOTA, 3.3\% on IDF1 and 14.9\% on FAF. Our method outperforms other 'private' models on Frag and ID.Sw. Compared with FairV1, the total number of ID.Sw drops from 6013 to 2237. Based on the performance on the three datasets, our tracker effectively optimizes tracking performance by preventing the performance degradation caused by missing detection.
	\begin{table}[t]
		\vskip 0.1in
		\begin{center}
			\begin{small}
				\setlength\tabcolsep{1pt}
				\fontsize{8}{10}\selectfont
				\begin{tabular}{c|c c  c c c c c}
					
					\hline
					
					Models & MOTA$\uparrow$ & IDF1$\uparrow$& MT$\uparrow$ & ML$\downarrow$ &FAF$\downarrow$& ID.Sw$\downarrow$ &Frag $\downarrow$ \\
					\hline
					\multicolumn{8}{c}{MOT2020 Test Dataset}\\
					\hline
					SORT\cite{bewley2016simple} &42.7
					&45.1
					&16.7$\%$
					&\color{red}{26.2$\%$}
					&{\color{red}6.1}
					&4470
					&17798\\
					SORT with CT    &{\color{red}43.3}
					&{\color{red}45.2}
					&{\color{red}17.6$\%$}
					&26.3$\%$
					&6.3
					&{\color{red}2971}
					&{\color{red}7485}\\
					\hline
					MOTDT\cite{chen2018real}*&43.7&40.8&14.4\%&27.7\%&\color{red}{3.4}&3705&9225\\
					MOTDT with CT&\color{red}{45.1}&\color{red}{41.9}&\color{red}{18.8\%}&\color{red}{27.1\%}&4.6&\color{red}{3494}&\color{red}{6712}\\
					\hline
					Tracktor\cite{bergmann2019tracking}& 52.6&\color{red}{52.6}&365&331&/&\color{red}{1648}&/
					\\
					Tracktor with CT &\color{red}{53.8}&51.1&\color{red}{386}&\color{red}{310}&2.0&2456&5046
					\\
					\hline
					\multicolumn{8}{c}{MOT2020 Training Dataset}\\
					\hline
					SORT\cite{bewley2016simple}*&45.8
					&34.1
					&288
					&593
					&/
					&12992
					&/
					\\
					SORT with CT     &{\color{red}52.9}
					&{\color{red}52.2}
					&{\color{red}424}
					&{\color{red}417}
					&/
					&{\color{red}3739}
					&/\\
					\hline
					MOTDT\cite{chen2018real}*&48.5&40.1&393&499&/&7754&/
					\\
					MOTDT with CT&\color{red}{50.2}&\color{red}{40.7}&\color{red}{498}&\color{red}{499}&/&\color{red}{7421}&/
					\\
					\hline
					Tracktor\cite{bergmann2019tracking}&66.4&\color{red}{60.7}&892&259&/&\color{red}{2664}&/
					\\
					Tracktor wtih CT &\color{red}{68.7}&60.6&\color{red}{926}&\color{red}{251}&/&3654&/
					\\
					\hline
				\end{tabular}
				
			\end{small}
			\caption{‘Public’ model ablation experiments. '*' means that the result is evaluated by ourselves.}
			\label{wacv-table:5}
		\end{center}
		\vskip -0.3in
	\end{table}
	\begin{table}[t]
		\vskip 0.1in
		\begin{center}
			\begin{small}
				\setlength\tabcolsep{1pt}
				\fontsize{8}{10}\selectfont
				\begin{tabular}{c|cccccc} 
					\hline
					
					Models & MOTA$\uparrow$ & IDF1$\uparrow$ & MOTP$\uparrow$ &MT$\uparrow$& ML$\downarrow$ & ID.Sw$\downarrow$  \\
					\hline
					\multicolumn{7}{c}{MOT2016 Test Dataset}\\
					\hline
					JDE\cite{wang2019towards} & 64.4&55.8& /&35.4$\%$ &20.0$\%$&1544\\
					POI\cite{yu2016poi}&66.1&65.1&79.5&34.0$\%$&20.8$\%$&\color{red}{805}\\
					Tube TK POI\cite{pang2020tubetk}&66.9&62.2&78.5&39.0$\%$&16.1$\%$&1236\\
					CTracker\cite{peng2020chained}&67.6&57.2&78.4&32.9$\%$&23.1$\%$&1897\\
					FairV1\cite{zhan2020simple}&68.7&70.4&{\color{red}80.2}&39.5$\%$&19.0$\%$&953\\
					
					QuasiDense\cite{pang2020quasi}&69.8&67.1&79.0&41.6\%&19.8\%&1097\\
					Ours&{\color{red}69.8}&{\color{red}71.1}&80.0&{\color{red}42.0$\%$}&{\color{red}15.8$\%$}&912
					\\
					\hline
					\multicolumn{7}{c}{MOT2017 Test Dataset}\\
					\hline
					SST\cite{sun2019deep}& 52.4
					& 49.5
					& 76.9
					& 21.4$\%$
					& 30.7$\%$
					&8431\\
					Tube\_TK\cite{pang2020tubetk}&63.0&58.6&78.3&31.2$\%$&19.9$\%$&4137\\
					CTracker\cite{peng2020chained} &66.6&57.4&78.2&32.2$\%$&24.2$\%$&5529\\
					FairV1\cite{zhan2020simple}&67.5&69.8&{\color{red}80.3}&37.7$\%$&20.8$\%$&2868\\
					CTTracker17\cite{zhou2020tracking}&67.8&64.7&78.4&34.6$\%$&24.6$\%$&3039\\
					QuasiDense\cite{pang2020quasi}&68.7&66.3&79.0&40.6\%&21.9\%&3378\\
					Ours&{\color{red}68.8}&{\color{red}70.2}&80.0&{\color{red}40.8$\%$}&{\color{red}17.7$\%$}&{\color{red}2805}\\
					
					\hline
				\end{tabular}
				
			\end{small}
			\caption{ Comparison experiment on MOT2016 and MOT2017.}
			\label{wacv-table:6}
		\end{center}
		\vskip -0.3in
	\end{table}
	\begin{table}[t]
		\vskip 0.10in
		\begin{center}
			\begin{small}
				\setlength\tabcolsep{1pt}
				\fontsize{8}{10}\selectfont
				\begin{tabular}{c|c c c c c c c c}
					\hline
					\multicolumn{9}{c}{MOT2020 Test Dataset}\\
					\hline
					Models & MOTA$\uparrow$ & IDF1$\uparrow$ & MOTP$\uparrow$ &MT$\uparrow$& ML$\downarrow$&FAF$\downarrow$ & ID.Sw$\downarrow$&Frag$\downarrow$  \\
					\hline
					FairV1\cite{zhan2020simple}& 58.7&63.7&77.2&{\color{red}66.3$\%$}&{\color{red}8.5$\%$}&24.7&6013&8140\\
					TrTrack\cite{sun2020transtrack}&64.5&59.2&\color{red}{80.0}&49.1\%&13.6\%
					&\color{red}{6.4}&3565&11383\\
					Ours&{\color{red}66.0}&{\color{red}67.0}&77.8& 56.3$\%$&13.3$\%$&9.8&\color{red}{2237}&{\color{red}4154}\\
					\hline
				\end{tabular}
				
			\end{small}
			\caption{Comparison experiment on MOT2020.}
			\label{wacv-table:7}
		\end{center}
		\vskip -0.35in
	\end{table}
	\section{Conclusion}
	In this paper, we point out the shortcomings of tracking by detection model and analyze the phenomenon of lost object in the real-time model in dense crowd area. Considering the computation burden, we propose a simple and effective compensation tracker and name it as CT. The proposed method has the advantage of plug and play without re-training the network. Our method employs tradition methods to re-track lost objects instead of additional networks. Extensive experiments indicate that the proposed method is able to effectively improve the tracking performance of real-time models with limited time consumption.

{\small
		\bibliographystyle{ieee_fullname}
		\bibliography{egbib.bib}
}

\end{document}